\def\year{2021}\relax
\documentclass[letterpaper]{article} % DO NOT CHANGE THIS
\usepackage{aaai21}  % DO NOT CHANGE THIS
\usepackage{times}  % DO NOT CHANGE THIS
\usepackage{helvet} % DO NOT CHANGE THIS
\usepackage{courier}  % DO NOT CHANGE THIS
\usepackage[hyphens]{url}  % DO NOT CHANGE THIS

\usepackage{graphicx} % DO NOT CHANGE THIS
\usepackage[ruled,vlined]{algorithm2e}
\usepackage{amsmath}
\usepackage{amssymb}
\usepackage{bm}
\usepackage{subfigure}
\usepackage{threeparttable}
\usepackage{multirow}
\usepackage{booktabs}
\usepackage[switch]{lineno}
\usepackage{natbib}  % DO NOT CHANGE THIS AND DO NOT ADD ANY OPTIONS TO IT
\usepackage{caption} % DO NOT CHANGE THIS AND DO NOT ADD ANY OPTIONS TO IT
\title{Neural Architecture Search as Sparse Supernet}

\author {
		Yan Wu\textsuperscript{\rm 1}\thanks{Equal contribution},
        Aoming Liu\textsuperscript{\rm 1}\footnotemark[1],
        Zhiwu Huang\textsuperscript{\rm 1},
        Siwei Zhang\textsuperscript{\rm 1},
        Luc Van Gool\textsuperscript{\rm 1,2}\\
}
\affiliations {
    \textsuperscript{\rm 1}Computer Vision Lab, ETH Z\"urich, Switzerland \\ \textsuperscript{\rm 2}PSI, KU Leuven, Belgium \\
    \{wuyan, aoliu\}@student.ethz.ch, siwei.zhang@inf.ethz.ch, \{zhiwu.huang, vangool\}@vision.ee.ethz.ch
}

\begin{document}
% \linenumbers
\maketitle

%\footnotetext[1]{Equal contribution}

\begin{abstract}
This paper aims at enlarging the problem of Neural Architecture Search (NAS) from Single-Path and Multi-Path Search to automated Mixed-Path Search. In particular, we model the NAS problem as a sparse supernet using a new continuous architecture representation with a mixture of sparsity constraints. The sparse supernet enables us to automatically achieve sparsely-mixed paths upon a compact set of nodes. To optimize the proposed sparse supernet, we exploit a hierarchical accelerated proximal gradient algorithm within a bi-level optimization framework. Extensive experiments on Convolutional Neural Network and Recurrent Neural Network search demonstrate that the proposed method is capable of searching for compact, general and powerful neural architectures.
\end{abstract}

\section{Introduction}
While deep learning has proven its superiority over manual feature engineering,
most of the conventional neural network architectures are still handcrafted by experts in a tedious and ad hoc fashion. Neural Architecture Search (NAS) has been suggested as the path forward for alleviating the network engineering pain by automatically optimizing architectures.
The automatically searched architectures perform competitively in computer vision tasks such as image classification ~\cite{zoph2016neural,liu2017hierarchical,zoph2018learning,liu2018darts,real2019regularized,chen2019renas,luo2018neural,cai2018proxylessnas,wu2019fbnet,zheng2019multinomial,you2020greedynas}, object detection \cite{zoph2018learning}, semantic segmentation \cite{liu2019auto,chen2019fasterseg} and image generation \cite{gong2019autogan,tian2020off}.

As one of the most popular NAS families, one-shot NAS generally models the architecture search problem as a one-shot training process of an over-parameterized supernet that comprises candidate architectures (paths). From the supernet, either Single-Path or Multi-Path architecture can be derived. However, both the existing Single-Path and Multi-Path Search works typically require a predefined structure on the searched architectures. For the Single-Path Search, some works like \cite{liu2018darts,liu2018progressive} search for a computation cell as the backbone block of the final architecture. Based on the modeling of directed acyclic graph, the cell comprises a set of nodes, each of which corresponds to a feature map, as well as their associated edges that represent single operations such as convolution and max-pooling. The resulting neural networks are limited to Single-Path architectures where each intermediate feature map is processed by a single operation. On the other hand, while Multi-Path architecture search methods like \cite{chu2020mixpath} search for a more flexible architecture with multiple paths between nodes, they generally require to fix the number of paths in advance. Moreover, both the existing Single-Path and Multi-Path Search methods have to manually fix the node number, which is another strong constraint for architecture search.
\begin{figure*}
  \centering
  %\vspace{-1em}
  \includegraphics[width=0.5\linewidth]{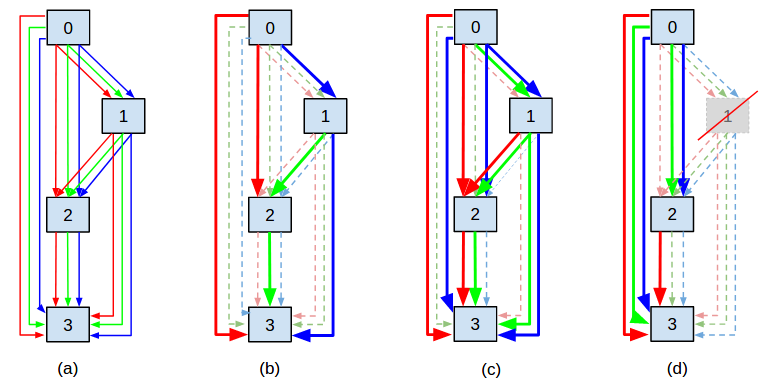}
%  \captionsetup{font={small}}
  \caption{Overview of various architecture search tasks over (a) Supernet that represents the continuous relaxation of the search space: (b) Single-Path Architecture Search that optimizes architectures with one single operation between each node pair, (c) Multi-Path Architecture Search that searches for multiple paths with a fixed amount between two nodes, and (d) the proposed Mixed-Path Architecture Search that has no rigid constraint on the path and node structure.}
  \label{fig:mixed-path}
%  \vspace{-1.2em}
\end{figure*}

In this paper, we target for a more automated NAS which can automatically optimize the mixture of paths as well as a changeable set of nodes. In other words, the target of automated Mixed-Path Architecture Search is to reduce unnecessary constraints on the structure of the searched architecture so as to explore in a more broad and general search space. For this purpose, we model the automated NAS problem as a one-shot searching process of a sparse supernet, which consists of sparsely-mixed paths and nodes without loss of network power. In particular, we exploit a new continuous architecture representation using Sparse Group Lasso to achieve the sparse supernet. As a result, the supernet is not only able to produce diverse mixed-paths between different pairs of nodes, but also automatically removes some useless nodes. The more general modeling however makes the optimization much more challenging due to the complex bi-level optimization with the non-differentiable sparsity constraint, which cannot be optimized by traditional network optimization algorithms well. To address this challenging issue, we propose a hierarchical accelerated proximal gradient algorithm that is capable of addressing the mixed sparsity under the bi-level optimization framework.
In summary, this paper brings several innovations to the domain of NAS as follows:
\begin{itemize}
    \item We suggest the new problem of Mixed-Path Neural Architecture Search where the node and path structure of the cell are automatically derived.
    \item We model the problem as a sparse supernet using a new continuous architecture representation with a mixture of sparsity constraints. 
    \item We propose a hierarchical accelerated proximal gradient algorithm to optimize the supernet search with the mixed sparsity constraints.
    \item We study that the searched Mixed-Path architectures are compact, general and powerful for some standard NAS benchmarks.
\end{itemize}

%%%%%%%%%%%%%%%%%%%%%%%%%%%%%%%%%%%%%%%%%%%%%%%%%%%%%%%%%%%%%%%%%%%%%%%%%%%%%%%%%%%%%%%%%%%%%%%%%%%%%%% To rearrange the problem statement and related work
\section{Problem Statement}
Neural Architecture Search aims at searching for computation cells as the building block of the final architecture. In general, each achitecture cell can be formulated as a directed acyclic graph (DAG) as shown in Fig.\ref{fig:mixed-path} (a), where each node represents a feature map in neural networks, and each directed edge is a mixture of operations that transform the tail node to the head node. As a consequence, the output of each intermediate node is a sum of incoming feature maps from predecessors. The DAG modeling enables the training of an over-parameterized supernet that stacks multiple basic cells. The optimal architecture is then derived from the supernet with the following three strategies.
\\
\\
\textbf{Single-Path Architecture Search} methods generally search for  one single operation between each node pair with fixed edge amounts in each cell as shown in Fig.\ref{fig:mixed-path} (b). To select a single path from a given supernet, they commonly first optimize the mixture of all associated operations, and finally choose those with the highest contribution to the supernet. As one of the most representative Single-Path Search methods, Differentiable Architecture Search (DARTS) \cite{liu2018darts} optimizes the mixture of operations within supernet using softmax combination. For each edge, the operation with largest softmax weight was selected. For each node, two input edges was selected by comparing each edge's largest operation weight. The rigid requirement on final architecture highly reduces the search space, such that the optimization process has a high potential to be stuck at a local minimum.
\\
\\
\textbf{Multi-Path Architecture Search} searches for multiple paths between any pair of nodes (Fig.\ref{fig:mixed-path} (c)), which is inspired by Multi-Path feature aggregations such as Inception networks \cite{szegedy2015going} and ResNeXt \cite{xie2017aggregated}. Nevertheless, the Multi-Path Search approach, such as \cite{chu2019fair, chu2020mixpath} typically still requires a strong prior knowledge about the aggregation intensity (i.e., path number) in advance of search. Furthermore, enforcing the same number of operations for each pair of nodes is very likely to reach a locally optimal architecture. 
\\
\\
\textbf{Mixed-Path Architecture Search} is hence proposed for the exploration of a more general search space to avoid the human intervene as much as possible. For this purpose, we suggest to enlarge the search space of NAS by relaxing the constraints on the network structure. In particular, the supernet is merely required to learn a complete and compact neural architecture, without any more rigid constraints on the node and path structure. In other words, it should be trained to automatically derive an optimal node and path structure as compact as possible without loss of classification or regression ability. The suggested new problem is conceptually illustrated in Fig.\ref{fig:mixed-path} (d). In comparison to Single-Path Search and Multi-Path Search, Mixed-Path Search dramatically increases the architecture search space, leading to a much more challenging NAS problem.

\section{Related Work}
\iffalse
\textbf{Network Pruning}
targets for reducing the model complexity by removing redundant network weights, neurons, layers, \textit{etc.}. Some network pruning works \cite{SSS, SGL-in-DL, li2020group, ye2018rethinking, liu2017learning, wen2016learning, NIPS2016_6372} proposed to impose sparsity constraint on network weights or auxiliary scale factors so as to sparsify the networks. In particular, \cite{SGL-in-DL} applies the Sparse Group Lasso constraint to remove network neurons, weights and select active input features. However, the search space of network pruning is fundamentally different from Neural Architecture Search. Network pruning focuses on pruning the neurons, weights or layers, while NAS is expected to focus on pruning the connections between different layers, namely structural connections. 
\fi
%\zhiwu{GreedyNAS...}
For \textbf{Neural Architecture Search}, early one-shot models generally aim for \textbf{Single-Path architecture search}~\cite{liu2018darts,cai2018proxylessnas, wu2019fbnet,liang2019darts+,xie2018snas,chen2019progressiveDARTS,li2019sgas}. For instance, DARTS \cite{liu2018darts} introduces a continuous relaxation of the discrete search space by aggregating candidate paths with softmax weights, so that a differentiable Single-Path Architecture Search can be performed. Based on DARTS, improvements including progressive search \cite{chen2019progressiveDARTS} 
and early stopping \cite{liang2019darts+} are proposed. In addition, ProxylessNAS \cite{cai2018proxylessnas} and FBNet \cite{wu2019fbnet} perform Single-Path Architecture Search with single-path sampling.

\textbf{Multi-Path Architecture Search} problem is proposed and addressed by some recent works.
For example, MixPath~\cite{chu2020mixpath} activates $m$ paths each time and a Shadow Batch Normalization is proposed to stabilize the training. GreedyNAS~\cite{you2020greedynas} is also Multi-Path architecture search with activating multiple paths. CoNAS~\cite{CoNAS} achieves Multi-Path Architecture Search by sampling sub-graph from a pre-trained one-shot model and doing Fourier analysis based on sub-graphs' performances. Multiple paths are selected based on the coefficients. Path amount of Fourier basis is controlled by its degree $d$. FairDARTS \cite{chu2019fair} uses sigmoid instead of softmax to eliminate the unfair optimization and allows multi paths, but it limits the maximum 2 paths between nodes. Generally, Multi-Path architecture search is still limited to search for a fixed number of paths.

Few works approach the variants of our defined \textbf{Mixed-Path Architecture Search}. For instance, DSO-NAS~\cite{DSO-NAS} enforces the $\ell1$-norm sparsity constraint (i.e., Lasso) to individual architecture parameters, which can achieve sparsely-mixed paths but overlooks the quest for sparse node structures especially when nodes are redundant initially (e.g., DARTS' cell structure). BayseNAS \cite{zhou2019bayesnas} exploits either $\ell1$-norm sparsity (with the same drawback with \cite{DSO-NAS}) or group-level sparsity with a weighted Group Lasso constraint in the classic Bayesian leaning manner, leading to sparse node structures. However, their Group Lasso constraint focuses on the sparsity on groups (i.e., nodes), and theoretically it may reach unsatisfactory sparsity on elements (i.e., paths). As a concurrent work, Gold-NAS~\cite{bi2020gold} suggests to gradually prune individual paths using one-level optimization to approach a mixed-path structure. By comparison, our Mixed-Path Architecture Search problem aims at both node and path structures. To this end, we model the Mixed-Path Architecture Search as a supernet with the Sparse Group Lasso constraint, which enables us to go for a more compact structure of nodes and paths. 
This enables our work to serve as a valuable pioneer study for such a more general Mixed-Path Architecture Search problem.
\\
\\
\textbf{Sparsity Constraints}, including Lasso \cite{lasso}, Group Lasso \cite{grouplasso}, Sparse Group Lasso \cite{sparse-group-lasso}, \textit{etc.} have been widely applied to areas such as statistics, machine learning and deep learning. A close application of sparsity constraints to NAS is the network pruning task which targets for reducing the model complexity by removing redundant network weights, neurons, layers, \textit{etc.}. Some network pruning works \cite{SSS, SGL-in-DL, li2020group, ye2018rethinking, liu2017learning, wen2016learning, NIPS2016_6372} proposed to impose sparsity constraints on network weights or auxiliary scale factors so as to sparsify the networks. In particular, \cite{SGL-in-DL} applies the same Sparse Group Lasso constraint to remove network neurons and weights. There are two major differences 
from our work: 1) The search space of \cite{SGL-in-DL} is fundamentally different from NAS. \cite{SGL-in-DL} imposes SGL constraint on the filter weights and focuses on pruning the neurons and weights, while ours focuses on pruning the connections between different layers, namely structural connections, and the sparsity constraint is imposed on architecture weights which yields a more challenging sparsity constrained problem under the bi-level optimization framework. 2) \cite{SGL-in-DL} merely adopts the traditional Adam to optimize the one-level Sparse Group Lasso constrained optimization problem. It is known that stochastic gradient descent algorithms, such as SGD and Adam are not proper to address non-differentiable optimization problem. Some works like \cite{sparse-group-lasso, fast_sgl} have learned blockwise descent algorithms to optimize Sparse Group Lasso, but they cannot be trivially applied to the stochastic optimization framework. To address this, we propose a hierarchical proximal bi-level optimization algorithm.

\section{Sparse Supernet}
Our Mixed-Path Architecture Search starts from an over-parameterized supernet, and aims at deriving an compact and optimal neural architecture. With the target of automatically selecting useful operations and nodes within the supernet, we are inspired by the prevailing sparsity regularization which can act as an automated feature selection mechanism. We thereby consider to introduce a sparse constraint to our supernet to select meaningful intermediate feature maps automatically. With the imposed sparsity constraint, we enable an automated sparse Mixed-Path Architecture Search.
\par
The supernet is designed as a stack of repetitive cells, and each cell is formulated as a DAG cell as shown in Fig.\ref{fig:mixed-path} (a). In particular, the mixture of operations on each edge is formulated in a ``regression-like" way. Instead of employing the widely-used softmax combination and its variants, such as Gumbel softmax~\cite{DATA}, we formulate the edge $e_{ij}$ between node $x_i$ and $x_j$ as a linear combination of operations, and the feature map derived from each operation $o \in O$ is scaled by a weight factor $A_{ij}^o$. The output feature map of intermediate node is now a scaled linear combination of various feature maps from different predecessors with their associated operations, i.e.,
\begin{linenomath}
\begin{align}
     x_j &= \sum_{i<j}\sum_{o \in O}A_{ij}^oo(x_i), A_{ij}^o \in \mathbb{R}^1
\end{align}
\end{linenomath}
\begin{algorithm*}
 \caption{Bi-level Optimization with the Proposed Hierarchical Accelerated Proximal Gradient (HAPG) Algorithm}
\SetAlgoLined
%\small
 {\textbf{Require}:
Supernet parameterized by $w$ and $A$ with $A_{i,j}^o$ being applied for each operation $o$ between nodes $i, j$};\\
 \While{not converged}{
  \textbf{Step1}: Update architecture weights A with HAPG given in Eq.\ref{hapg1}-\ref{prox2}. Note that the gradient are computed with the second order approximation given in  Eq.\ref{second_order}.\\
  \textbf{Step2}: Update $w$ by descending $\nabla_wl_{train}(w, A)$;
 }
 \textbf{Ensure}: Sparse supernet based on sparse architecture weights A.
 \label{alg:alg1}
\end{algorithm*}

To relax the structure constraints on both the number of nodes and paths per edge, we aim at achieving the operation sparsity as well as the node sparsity. Sparse Group Lasso (SGL) regularization \cite{sparse-group-lasso} meets our expectation exactly, which allows for both element (operation) sparsity and group (node) sparsity. In a DAG with $N$ intermediate nodes, for each node $x_j$, we group weight factors for all incoming feature maps $A_{ij}^o$, where $i<j, o \in O$ as $A_{(j)}$. Mathematically, the full objective function is derived as:
% \begin{small}
\begin{linenomath}
\begin{align}
    &L(w, A) %&= l(w, A) + \Omega(A)\\
    =l(w, A) + \Omega_{SGL}(A) \\
    &= l(w, A) + \lambda\alpha||A||_1 + \lambda(1-\alpha)\sum_{n=1}^N\sqrt{|A_{(n)}|}\cdot||A_{(n)}||_2 \label{target_function}
\end{align}
\end{linenomath}
% \end{small}
\noindent where $\Omega_{SGL}$ corresponds to the mixed sparsity regularization, $\lambda$ is the sparsity strength factor, and $\alpha$ controls the balance between operation sparsity and node sparsity. By optimizing the network parameters $w$ and the architecture weights $A$ with the target function in Eq.\ref{target_function}, we can achieve a sparse supernet structure.
Ideally, the final architecture will be derived by removing the operations with zero weights and the nodes with all zero incoming weights.

To jointly optimize the supernet and learn a sparse network structure, we target for solving the following bi-level optimization problem:
\begin{linenomath}
\begin{align}
    \min_A \hspace{0.3cm}&l_{val}(w^*(A), A) + \Omega_{SGL}(A) \nonumber\\
    s.t. \hspace{0.3cm} &w^*(A) = \textrm{argmin}_w \hspace{0.15cm}l_{train}(w, A)
\end{align}
\end{linenomath}
where the network weights $w$ and the architecture weights $A$ are optimized on two separate training and validation sets to avoid architecture from overfitting to data.

\section{Optimization}
As $\ell_1$-norm term is convex but non-differentiable, the SGL regularization term yields a challenging optimization problem. Conventional stochastic gradient descent algorithms, such as SGD and Adam generally cannot work well. While some exiting works like \cite{sparse-group-lasso, fast_sgl} have exploited blockwise descent algorithms to fit SGL, it is non-trivial to apply their algorithms to the stochastic optimization setting. We thereby turn to the proximal methods \cite{proximal_optimization} which is capable of solving the optimization problem with the non-differentiable term and enables us to learn some exact zero weights via soft-threshold. We propose a Hierarchical Accelerated Proximal Gradient algorithm (\textbf{HAPG}) and its improved version (\textbf{AdamHAPG}), both of which are suitable for stochastic optimization. Finally, we further appropriately incorporate these two methods into the bi-level optimization framework.

\begin{table*}[t]
\centering
%\small
%\setlength{\abovecaptionskip}{5pt} 
%\setlength{\belowcaptionskip}{-5pt}
%\captionsetup{font={small}}
 \begin{threeparttable}[b]
%  \resizebox{\textwidth}{!}{
\begin{tabular}{@{}lccccccc@{}}
\toprule
\multirow{2}{*}{\textbf{Architecture}} & \multicolumn{2}{c}{\textbf{Test Error (\%)}}  & \textbf{Params} & \textbf{Search Cost} & \textbf{Architecture} \\
\cline{2-3}
                                           & \textbf{C10}      & \textbf{C100}      & \textbf{(M)} & \textbf{(GPU days)}  & \textbf{Type} \\ 
\midrule
DenseNet-BC~\cite{DenseNet} & 3.46 & 17.18 & 25.6 & -- & manual \\
\midrule
DARTS (first order)~\cite{liu2018darts}                             & 3.00 $\pm$ 0.14 & 17.76 & 3.3 & 1.5 & Single-Path \\
DARTS (second order)~\cite{liu2018darts}                          & 2.76 $\pm$ 0.09 & 17.54 &  3.3  & 4 & Single-Path \\  
P-DARTS \cite{chen2019progressiveDARTS}                          & 2.50 & 16.55 &  3.4  & 0.3 & Single-Path \\  
PC-DARTS \cite{xu2020pc-darts} & 2.57 $\pm$ 0.07 & -- &  3.6  & 0.1 & Single-Path \\  
%DARTS+$^\ddagger$  \cite{liang2019darts+}                        & 2.50 $\pm$ 0.11 & 16.28 &  3.4  & 0.3 & Single-Path \\
%MixPath-c~\cite{chu2020mixpath} & 2.60 & -- & 5.4 & 0.25  & Multi-Path\\
FairDARTS~\cite{chu2019fair}& 2.54 $\pm$ 0.05 & -- & 3.32 $\pm$ 0.46  & 0.5  & Multi-Path \\
CoNAS~\cite{CoNAS} & 2.62 $\pm$ 0.06 & -- & 4.8 & 0.7  & Multi-Path\\
DSO-NAS~\cite{DSO-NAS}& 2.84 $\pm$ 0.07 &-- & 3.0  & 1  & Mixed-Path \\
%DSO-NAS-full\cite{DSO-NAS} & 2.95 $\pm$ 0.12 & -- & 3.0  & 1  & mixed-path \\
BayesNAS~\cite{zhou2019bayesnas}&2.81 $\pm$ 0.04 & -- & 3.4 & 0.2 & Mixed-Path\\

\midrule
SparseNAS + HAPG  & 2.73 $\pm$ 0.05 & 16.83 & 3.8  & 1 & Mixed-Path \\
SparseNAS + AdamHAPG   & 2.69 $\pm$ 0.03 & 17.04 &  4.2  & 1 & Mixed-Path \\
SparseNAS + AdamHAPG*   & 2.50 & 16.79 &  3.5  & 0.27 & Mixed-Path \\
\bottomrule
\end{tabular}
% }
\begin{tablenotes}
\tiny{
\item[*] {\footnotesize Obtained by searching in a cherry-picked search space.}
}
\end{tablenotes}
\end{threeparttable}
\caption{Performance Comparison on CIFAR-10 and the Transferability to CIFAR-100 (lower error rate is better).}
\label{tab:cifar-results}
\end{table*}

\begin{table*}[t]
	\centering
	\begin{threeparttable}[b]
%  \resizebox{\textwidth}{!}{
	\begin{tabular}{@{}lccccccc@{}}
		\toprule
		\multirow{2}{*}{\textbf{Architecture}} & \multicolumn{2}{c}{\textbf{Test Error (\%)}}      & \textbf{Params} & \textbf{Architecture} \\ \cline{2-3}
		& top-1 & top-5      & \textbf{(M)} &  \textbf{Type} \\ \midrule
		MobileNet \citep{mobilenet} & 29.40 & 10.5 & 4.2 & manual \\ 
		\midrule
		
		DARTS (second order)~\cite{liu2018darts} &  26.70   &  8.7 & 4.7 & Single-Path \\
        FairDARTS-B~\cite{chu2019fair} & 24.90 & 7.5 & 4.8 &  Multi-Path\\
		DSO-NAS~\cite{DSO-NAS} & 26.20 & 8.6 & 4.7 &  Mixed-Path\\
		BayesNAS~\cite{zhou2019bayesnas} & 26.50 & 8.9 & 3.9 & Mixed-Path\\
		\midrule
		SparseNAS + HAPG   &  25.48  &  8.1 & 5.3 & Mixed-Path \\
		SparseNAS + AdamHAPG    &  24.67  &  7.6 & 5.7 & Mixed-Path \\
		\bottomrule
	\end{tabular}
\end{threeparttable}
	\caption{Transferability Comparison on ImageNet in the Mobile Setting (lower error rate is better)}
	\label{tab:imagenet-results}
\end{table*}
\subsection{Hierarchical Proximal Optimization}
Computing the proximal operator $\textrm{Prox}_\Omega(\cdot)$ of the regularization term $\Omega$ is a key part of proximal optimization algorithms. The joint combination of $\ell_1$ and $\ell_1/\ell_2$ norm in SGL brings much higher complexity to the direct proximal operator computing. Inspired by \cite{proximal_optimization} that the SGL norm is a special case of hierarchical norm \cite{hierarchical_norm}, with $\ell_1$-norm of each individual weight factor being a child group of the $\ell_1/\ell_2$-norm, we derive the hierarchical proximal operator as a composition of $\ell_1$-norm and $\ell_1/\ell_2$-norm proximal operators:
\begin{linenomath}
\begin{align}
    \textrm{Prox}_{\Omega} (\cdot) = \textrm{Prox}_{\lambda(1-\alpha)||\cdot||_2}  \circ  \textrm{Prox}_{\lambda\alpha||\cdot||_1} (\cdot)
\end{align}
\end{linenomath}

As for proximal algorithms, widely-used methods include ISTA and FISTA \cite{ISTA}, and here we employ an efficiently reformulated Accelerated Proximal Gradient (APG) optimization scheme \cite{SSS} which allows for the stochastic optimization setting. Accordingly, we propose a Hierarchical Accelerated Proximal Gradient (\textbf{HAPG}) algorithm tailored for the Sparse Group Lasso regularization:
%\vspace{-0.1cm}
\begin{linenomath}
\begin{align}
    z_t &= A_{t-1} - \eta_tg_{t-1} \label{hapg1}\\ 
    v_t &= \textrm{Prox}_{\eta_t\lambda(1-\alpha)||\cdot||_2}  \circ  \textrm{Prox}_{\eta_t\lambda\alpha||\cdot||_1}(z_t) \nonumber\\ 
    & - A_{t-1} + u_{t-1}v_{t-1} \label{hapg2}\\
    A_t &= \textrm{Prox}_{\eta_t\lambda(1-\alpha)||\cdot||_2}  \circ  \textrm{Prox}_{\eta_t\lambda\alpha||\cdot||_1}(z_t) + u_tv_t \label{hapg3}
\end{align}
\end{linenomath}
where $g_{t-1}$ represents the gradient, $\eta_t$ is the gradient step size and $u_t = \frac{t-2}{t+1}$. And the proximal operators can be derived as: 
\small
\begin{linenomath}
\begin{align}
    [\textrm{Prox}_{\eta_t\lambda\alpha||\cdot||_1}(\mathbf{z})]_i &= \textrm{sgn}(z_i)(|z_i|-\eta_t\lambda\alpha)_+ \label{prox1}\\
    [\textrm{Prox}_{\eta_t\lambda(1-\alpha)||\cdot||_2}(\mathbf{z})]_n &= \left(1 - \frac{\sqrt{|\mathbf{z_{(n)}}|}\eta_t\lambda(1-\alpha)}{||\mathbf{z_{(n)}}||_2}\right)_+ \mathbf{z_{(n)}} \label{prox2}
\end{align}
\end{linenomath}
\normalsize

To further facilitate the optimization, we introduce the powerful Adam into the proposed HAPG (\textbf{AdamHAPG}) and replace the gradient descent in Eq.\ref{hapg1} with an Adam gradient update \cite{adam}. We should note that each weight factor gets an individual gradient step size in AdamHAPG, and we thereby make small adaptations when computing proximal operators. As for the $\ell_1$-norm proximal operator, we implement the proximal update using the corresponding step size for each weight, while for the $\ell_1/\ell_2$-norm proximal operator, we heuristically take the median value of step sizes for each group as an approximation and we experimentally show that it works properly for our problem.

\subsection{Bi-level Optimization with Hierarchical Proximal Optimization}
We incorporate our proposed hierarchical proximal algorithms into the bi-level optimization framework \cite{liu2018darts} to alternatively optimize the network parameter $\omega$ and the architecture weight $A$.
In particular, we follow \cite{liu2018darts} to compute the gradient of the architecture weights (i.e., $g_{t-1}$ in Eq.\ref{hapg1}):
\small
\begin{linenomath}
\begin{align}
    g_t &= \nabla_Al_{val}(w^*(A_t), A_t) \label{second_1} \\
    &\approx \nabla_A l_{val}(w_t', A_t) \nonumber \\
    &- \gamma \nabla_{A, w}^2l_{train}(w_t, A_t)\nabla_{w'}l_{val}(w'_t, A_t) \label{second_2} \\
    &\approx \nabla_A l_{val}(w'_t, A_t) \nonumber \\
    &- \frac{\nabla_Al_{train}(w_t^+, A_t) - \nabla_Al_{train}(w_t^-, A_t)}{2\epsilon} \label{second_order}
\end{align}
\end{linenomath}
\normalsize
where $w'_t = w_t-\gamma\nabla_wl_{train}(w_t, A_t)$, $w_t^{\pm} = w_t \pm \epsilon\nabla_wl_{train}(w'_t, A_t)$ and $\epsilon$ and $\gamma$ are set to be small scalars as done in \cite{liu2018darts}. Eq.\ref{second_2} is derived by a one-step forward approximation, i.e., $w^*(A_t) \approx w'_t = w_t-\gamma\nabla_wl_{train}(w_t, A_t)$, and Eq.\ref{second_order} follows the second-order approximation in \cite{liu2018darts}. Especially, when introducing the HAPG and AdamHAPG to the bi-level optimization framework, to stabilize the training, we follow \cite{sparse-group-lasso} to adopt a similar pathwise solution for an incremental increase of regularization factor $\lambda$, and we experimentally show the effectiveness of this progressive sparsifying solution. The complete optimization 
algorithm is presented in Alg.\ref{alg:alg1}.

\section{Evaluation}

\begin{table*}[t]
\centering
%\small
%\setlength{\abovecaptionskip}{5pt} 
%\setlength{\belowcaptionskip}{-5pt}
%\captionsetup{font={small}}
%  \resizebox{\textwidth}{!}{
\begin{tabular}{@{}lccccccc@{}}
\toprule
\multirow{2}{*}{\textbf{Architecture}} & \multicolumn{2}{c}{\textbf{Perplexity}}  & \textbf{Params} & \textbf{Search Cost} & \textbf{Architecture} \\
\cline{2-3}
                                           & \textbf{valid}      & \textbf{test}      & \textbf{(M)} & \textbf{(GPU days)}  & \textbf{Type} \\ 
\midrule
LSTM \cite{merity2017regularizing} & 60.70 & 58.80 & 24 & -- & manual \\
\midrule
DARTS (first order)~\cite{liu2018darts}                             & 60.20 & 57.60 & 23 & 0.5 & Single-Path \\
DARTS (second order)~\cite{liu2018darts}                          & 58.10 & 55.70 &  23  & 1 & Single-Path \\  
CoNAS~\cite{CoNAS} & 59.10 & 56.80 & 23 & 0.25  & Multi-Path\\
\midrule
SparseNAS + AdamHAPG   & 57.73 & 55.37 &  23  & 0.25 & Mixed-Path \\
\bottomrule
\end{tabular}
\caption{Performance Comparison on PTB (lower error rate is better)}
\label{tab:ptb-results}
\end{table*}

\iffalse
\begin{table*}[t]
\centering
%\small
%\setlength{\abovecaptionskip}{5pt} 
%\setlength{\belowcaptionskip}{-5pt}
%\captionsetup{font={small}}
\caption{Performance Comparison on WikiText-2 (lower error rate is better)}
\label{tab:wt2-results}
 \begin{threeparttable}[b]
%  \resizebox{\textwidth}{!}{
\begin{tabular}{@{}lccccccc@{}}
\toprule
\multirow{2}{*}{\textbf{Architecture}} & \multicolumn{2}{c}{\textbf{Perplexity}}  & \textbf{Params} & \textbf{Search Cost} & \textbf{Architecture} \\
\cline{2-3}
                                           & \textbf{valid}      & \textbf{test}      & \textbf{(M)} & \textbf{(GPU days)}  & \textbf{Type} \\ 
\midrule
LSTM \cite{merity2017regularizing} & 69.10 & 66.00 & 33 & -- & manual \\
\midrule
%DARTS (first order)\cite{liu2018darts}                             & 60.20 & 57.60 & 23 & 0.5 & Single-Path \\
DARTS~\cite{liu2018darts}                          & 71.20 & 69.60 &  33  & 1 & Single-Path \\  
\midrule
SparseNAS + AdamHAPG   & -- & -- &  33  & 0.25 & Mixed-Path \\
\bottomrule
\end{tabular}
\end{threeparttable}
\end{table*}
\fi

\begin{figure*}
    \centering
  \includegraphics[width=15cm]{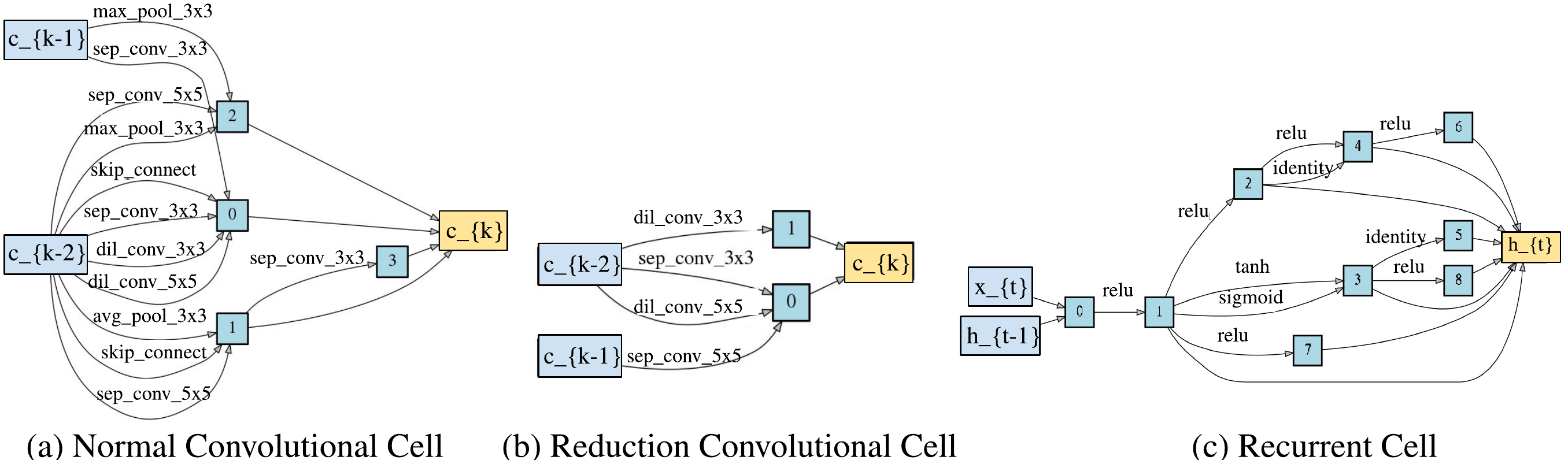}
%\captionsetup{font={small}}
    \caption{Architectures searched on CIFAR-10 and PTB}
    \label{fig:cells}
\end{figure*}

We evaluate the proposed SparseNAS for Convolutional Neural Network (CNN) and Recurrent Neural Network (RNN) architecture search on CIFAR-10 and Penn Treebank (PTB) respectively, and further investigate the transferability of searched architectures on CIFAR-10 to CIFAR-100 and ImageNet. In both CNN and RNN cell search experiments, we follow the setup of DARTS \cite{liu2018darts} to implement SparseNAS, where we use the same search space, cell setup and we stack the same number of cells for fair comparison. While there exist many direct improvements over DARTS like progressive search \cite{chen2019progressiveDARTS} and early stopping \cite{liang2019darts+}, most of them make orthogonal contributions to our work and thus can be used to improve our method as well. However, further applying them to our method is beyond the scope of our paper. Hence, we mainly take the most related Single-Path (DARTS~\cite{liu2018darts}), Multi-Path (FairDARTS~\cite{chu2019fair}, CoNAS~\cite{CoNAS}) and Mixed-Path (DSO-NAS~\cite{DSO-NAS}, BayesNAS~\cite{zhou2019bayesnas}) methods as our real competitors. Note that the reported results of all the competitors are from their original papers. For more detailed experiments setup, please refer to the Appendix. 

\subsection{Convolutional Neural Architecture Search} 
The convolutional cell is directly searched on CIFAR-10 and transferred to CIFAR-100 and ImageNet.

On CIFAR-10, HAPG method obtains architecture with error $2.73 \pm 0.05$. Due to the use of the adaptive learning rate, AdamHAPG ($2.69 \pm 0.03$) performs better. Both the HAPG and AdamHAPG based SparseNAS
outperform the second-order DARTS ($2.76 \pm 0.09$) and all Mixed-Path competitors. In particular, a search performed on a cherry-picked search space achieves the best performance (2.50) among all competitors.
The results are shown in Table \ref{tab:cifar-results}. 

The transferability results from CIFAR-10 to CIFAR-100 and ImageNet are shown in Table \ref{tab:cifar-results} and Table \ref{tab:imagenet-results} respectively. Note that some competitors like CoNAS did not report their transferability results from CIFAR-10 to CIFAR-100 and ImageNet. The reported results show that our  AdamHAPG performs the best (even better than the recent FairDARTS on ImageNet), and both the HAPG and AdamHAPG based SparseNAS outperform the other competitors. 

Note that we search for a sparse model structure, but not necessarily a model with small model size, and our method show a clear advantage in structural sparsity. We present the normal and reduction cell searched on CIFAR-10 in Figure \ref{fig:cells}(a) and (b). Notably, we get compact reduction cells with only 2 remained nodes, which proves our advantage in group-level sparsity compared with DOS-NAS \cite{DSO-NAS}. Compared with another Mixed-Path work BayesNAS \cite{zhou2019bayesnas}, we have fewer paths in normal cells and reduction cells and thus we achieve better element-level sparsity and even higher performance. Compared with the competitors, our searched architectures show more general properties with more diverse path and node structures. In addition, there is no obvious "collapse" problem (e.g. excessive skip-connection selected) problem observed.

\subsection{Reccurent Neural Architecture Search} Recurrent cell search is performed on the PTB dataset, and we follow the experiment setup of \cite{liu2018darts}. 
When optimizing with HAPG, we observe that the search phase does not converge with an exploded gradient of architecture weights which is a typical RNN training problem. The AdamHAPG with adaptive learning rate is shown to be helpful to stabilize the training for RNN model search.
The results are summarized in Table \ref{tab:ptb-results}. Except for DARTS and CoNAS, the other competitors did not report their results on PTB. To our best knowledge, the derived architecture by our method obtains a new state-of-the-art NAS on PTB with valid perplexity 57.73 and test perplexity 55.37. 

In Figure \ref{fig:cells}(c), we present the derived recurrent cell on PTB. Compared to DARTS and CoNAS where each node only receives a fixed number of  incoming edges, the cell derived by SparseNAS is more general that intermediate nodes can have different numbers of incoming edges. And this general architecture has shown its superior performance.

\iffalse
\begin{figure*}[htb]
\makeatletter\def\@captype{table}\makeatother
\begin{minipage}{0.6\textwidth}
\centering
 %   \small
    \begin{tabular}{cc}
    \toprule
         & \textbf{Test Error (\%)} \\
       \midrule
       Adam($\alpha=0.5$) & 2.93 \\
       \midrule
       AdamHAPG($\alpha$=0.3)  & 3.30 \\
       AdamHAPG($\alpha$=0.5)  & 2.69 \\
       AdamHAPG($\alpha$=0.7)  & 2.79 \\
       \bottomrule
    \end{tabular}
%    \captionsetup{font={small}}
        \caption{Comparison between Adam and AdamHAPG and ablation study on $\alpha$ on CIFAR-10}
    \label{tab:alpha_ablation} 
\end{minipage}
\makeatletter\def\@captype{figure}\makeatother
\begin{minipage}{0.4\textwidth}
\includegraphics[width=6cm]{val_acc.pdf}
%\captionsetup{font={small}}
    \caption{Valid accuracy during search stage with different $\lambda$ steps}
    \label{fig:step_ablation}
\end{minipage}
\end{figure*}
\fi

\begin{figure*}
    \centering
%    \captionsetup{font={small}}
    \includegraphics[width=15cm]{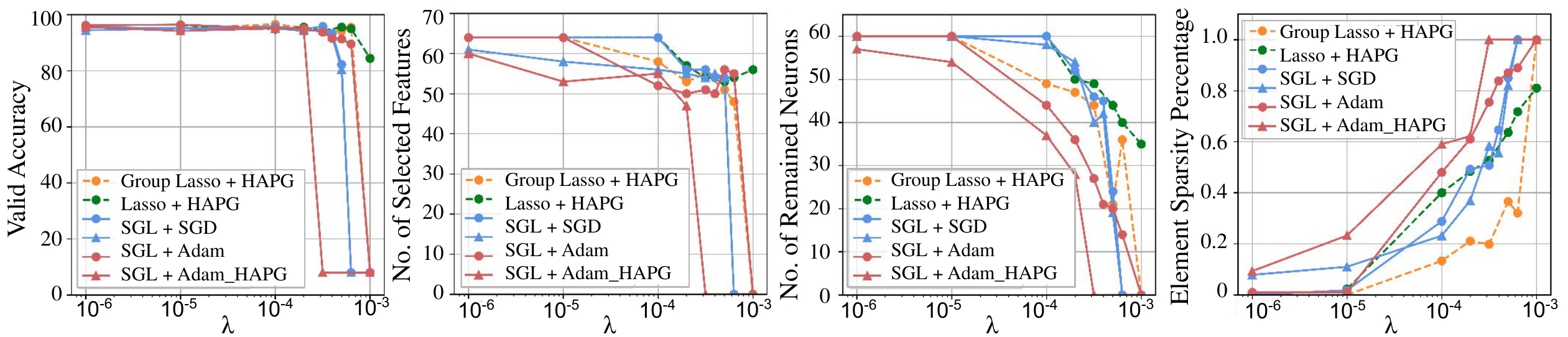}
    \caption{Comparison between Lasso, Group Lasso and Sparse Group Lasso, and comparison between different optimization methods, including SGD, Adam, HAPG and AdamHAPG. From left to right: the valid accuracy of standalone sparse network, the number of selected input features, the number of remained inner neurons, the total sparsity percentage of network weights. With comparable valid accuracy, the HAPG and AdamHAPG with Sparse Group Lasso give better element and group sparsity.}
    \label{fig:alpha_study}
    %\vspace{-1.2em}
\end{figure*}

\subsection{Ablation Study}
%\textbf{HAPG/AdamHAPG vs. SGD/Adam}
\textbf{HAPG/AdamHAPG vs. SGD/Adam} We study the advantage of our HAPG and AdamHAPG over conventional SGD and Adam in the sparsity constrained optimization problem. Following \cite{SGL-in-DL} that applies SGL constraint to network weights to select features and remove redundant weights and neurons, we conduct various experiments in this easier network pruning task, which is essentially a one-level sparsity constrained optimization problem, to purely evaluate the effectiveness and advantage of our proposed optimization algorithms. Starting with a fully connected network with two hidden layers (40 and 20 hidden neurons respectively), we implement classification task on DIGITS \cite{digits}. The 8$\times$8 images are flatten into 64-dim vectors as the input features.

The performances of these four methods are presented in Fig.\ref{fig:alpha_study}. From left to right, Fig.\ref{fig:alpha_study} shows the valid accuracy of stand-alone sparse network, number of selected features, number of remained inner neurons and the element sparsity percentage of network weights. The horizontal axis indicates the different sparsity constraint factor $\lambda$. As $\lambda$ increasing, stronger sparsity regularization derives sparser network. With $\lambda$ ranging from $10^{-5}$ to $10^{-3.7}$, stand-alone architectures are all well-performed with comparable performances. Whereas, in terms of sparsity, our proposed HAPG and AdamHAPG clearly outperform their counterparts Adam and SGD. In particular, AdamHAPG shows a clear superiority to have a more compact structure and more powerful feature selection without loss in the classification accuracy.

\textbf{SGL vs. Lasso and Group Lasso (GL)} 
With the same optimization method HAPG, 3 experiments with Lasso, GL, SGL constraints are conducted respectively. Fig.\ref{fig:alpha_study} shows that enforcing SGL is clearly better than using Lasso and GL both in group (features and neurons) and element (network weights) sparsity,  while having comparable valid accuracies.

\textbf{Effect of \bm{$\alpha$}} In Table \ref{tab:alpha_ablation}, we study the effect of different $\alpha$ on CNN task. Theoretically, $\alpha$ controls the balance between path sparsity and node sparsity. With $\alpha$ decreasing, the algorithm tend to obtain a more node-sparse architecture. Empirically, in CNN task, with AdamHAPG, the $\alpha=0.5$ obtains the optimal architecture with highest accuracy.

\textbf{Effect of \bm{$\lambda$}}  $\lambda$ is another hyperparameter to control the sparsity strength. Since we progressively increase the sparsity strength during search via a linearly increased $\lambda$, we study the effect of $\lambda$ increasing step to the training stability of search process. We plot the evolution of valid accuracy during CNN cell search phase with increasing $\lambda$ step 0.01, 0.02 and 0.03 respectively in Fig. \ref{fig:step_ablation}. With larger $\lambda$, we expect a better sparsity, but it is likely that architecture weights fluctuate heavily at each updating step, and this can explain that with step size 0.03, we see a large fluctuation in valid accuracy after 15 epochs. And small step sizes 0.01, 0.02 lead to a more stable training. As a trade-off between architecture sparsity level and stability of searching process, we typically choose a value like 0.01 as the step size in our experiments.
\begin{table}
% \small
    \centering
    \begin{tabular}{cc}
    \toprule
         & \textbf{Test Error (\%)} \\
 %      \midrule
%       Adam($\alpha=0.5$) & 2.93 \\
       \midrule
       AdamHAPG($\alpha$=0.3)  & 3.30 \\
       AdamHAPG($\alpha$=0.5)  & 2.69 \\
       AdamHAPG($\alpha$=0.7)  & 2.79 \\
       \bottomrule
    \end{tabular}
%    \captionsetup{font={small}}
    \caption{Study on the effect of $\alpha$ on CIFAR-10}
    \label{tab:alpha_ablation}
\end{table}

\begin{figure}
    \centering
  \includegraphics[width=4.5cm]{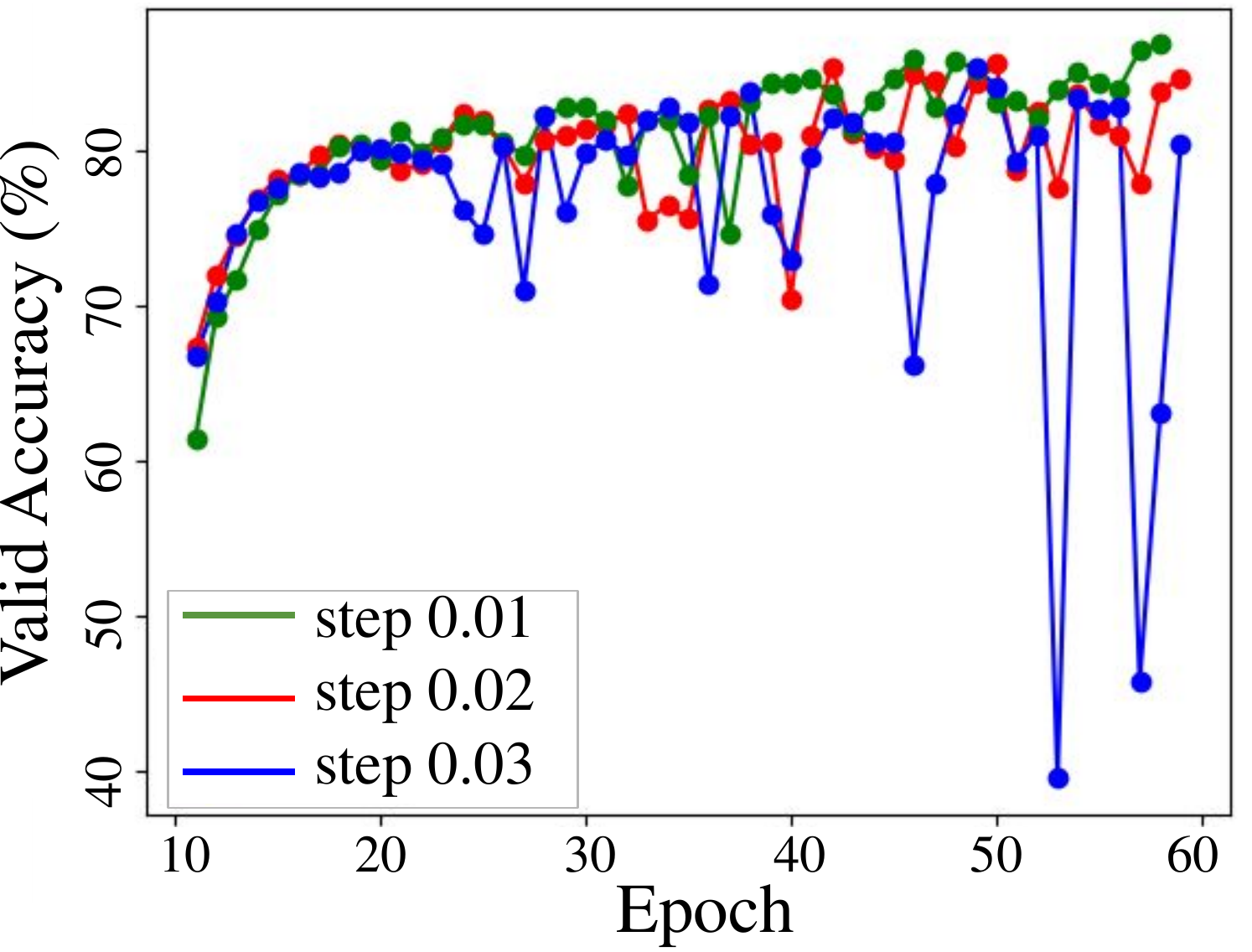}
%\captionsetup{font={small}}
    \caption{Valid accuracy during search with different $\lambda$ steps}
    \label{fig:step_ablation}
    \vspace{-0.3cm}
\end{figure}

\section{Conclusion}
In this work, we launch Neural Architecture Search to explore in a more general and flexible Mixed-Path Search space using a sparse supernet. Starting from a supernet parameterized by architecture weight factors, we exploit the Sparse Group Lasso regularization on weight factors to automatically search for optimal structures of nodes and paths. To address the challenging optimization problem with non-differentiable sparsity constraint, we propose novel hierarchical proximal algorithms and incorporate them into a bi-level optimization framework. We experimentally show very competitive results and potentials of our derived Mixed-Path architectures on various datasets. We believe that our general Mixed-Path Search modeling will lead the future NAS research to a much broader search space and bring the possibility to derive more flexible and powerful architectures.

\section*{Acknowledgements}
This work was supported by the ETH Z\"urich Fund (OK), an Amazon AWS grant, and an Nvidia GPU grant.

\section*{Appendix}
\appendix
\section{Experimental Setting}

% \subsection{Neural Architecture Search}
\subsection{Convolutional Neural Architecture Search}
In the convolutional architecture search stage, we optimize the supernet for 50 epochs with batch size 64. The supernet is a stack of 8 basic cells and each cell has 7 nodes with 2 inputs and one single output node. The channel number is set to be 16. To optimize the network parameters $w$, we employ SGD momentum with momentum 0.9, weight decay $3\times10^{-4}$ and initiate the learning rate as 0.025 (an annealing cosine decay schedule is applied). To optimize the architecture weight factors $A$, we employ the proposed HAPG and AdamHAPG. For HAPG, we initiate the learning rate as 0.025 with an annealing cosine decay schedule. For AdamHAPG, the initial learning rate is $3\times10^{-4}$ and for the Adam update step, the momentum is (0.5, 0.999). 
The search space is almost the same as that of DARTS \cite{liu2018darts} (except for the use of the \textit{zero} operation), which includes \textit{skip-connect}, \textit{max-pool-3$\times$3}, \textit{avg-pool-3$\times$3}, \textit{sep-conv-3$\times$3}, \textit{sep-conv-5$\times$5}, \textit{dil-conv-3$\times$3}, \textit{dil-conv-5$\times$5}.
We also conduct experiments on a cherry-picked search space with optional candidates \textit{sep-conv-3$\times$3}, \textit{skip-connect} for normal cell and \textit{skip-connect}, \textit{max-pool-3$\times$3} for reduce cell.
\\
\\
\iffalse
\textbf{Cherry-picked Search Space} Mixed-Path architecture search without any constraints explores in a huge search space ($1\times10^{57}$). Thus we design a small search space comparable with DARTS to show Mixed-Path architecture's advantage and potential.
We shrink the search space with only \textit{sep-conv-3$\times$3}, \textit{skip-connect} for normal cell and \textit{skip-connect}, \textit{max-pool-3$\times$3} for reduce cell. The search space scale with Mix-Path Architecture is approximately $1\times10^{17}$, while DARTS is approximately $1\times10^{18}$. Other settings are similar as before. 
\fi
The architecture searched on CIFAR-10 is evaluated on CIFAR-10 and transfered to CIFAR-100 and ImageNet. As CIFAR-100 is highly related to CIFAR-10, we do not do direct architecture search on CIFAR-100 redundantly. Following DARTS, we stack the extracted cells for 20 times and use 36 initial channels for evaluation network. As the architecture topology from Mix-Path search is relatively complex, we train the network for 1200 epochs to ensure convergence. we also add cutout, path dropout with probability 0.2 and auxiliary towers with weight 0.4. Meanwhile, the architecture from small search space is trained with the same setting as DARTS. For ImageNet mobile setting, the size of input images is 224$\times$224. The cell is stacked for 14 times an trained for 250 epochs with batch size 512, weight decay $3\times10^{-5}$, SGD optimizer and linear learning decay with initial learning rate to be 0.4.

\begin{figure*}
	\centering
	\vspace{-1em}
	\includegraphics[width=0.9\textwidth]{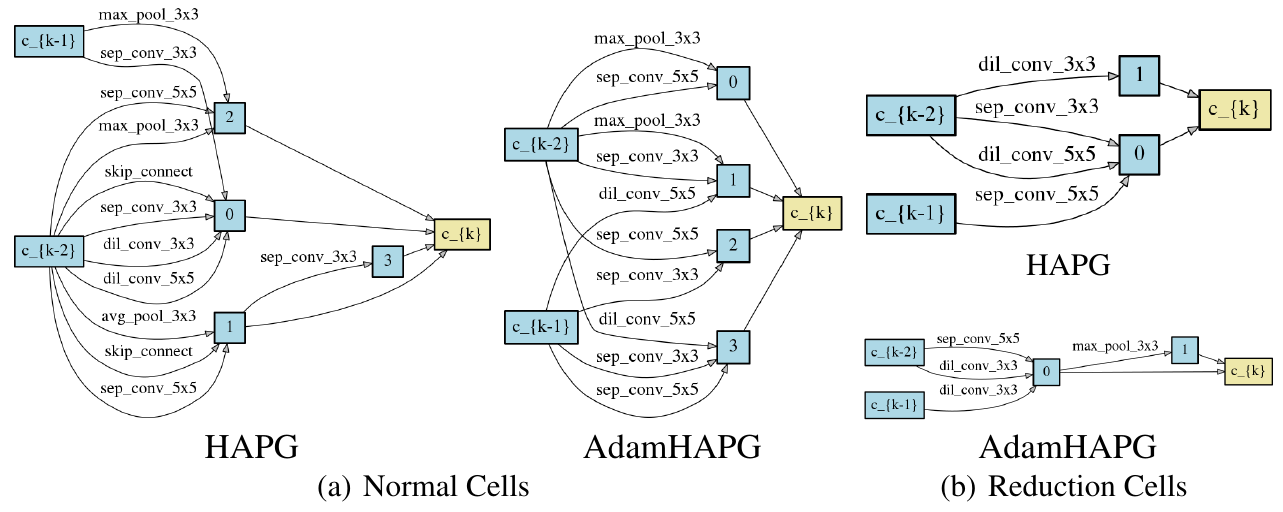}
	%\captionsetup{font={small}}
	\caption{Convolutional cell searched on CIFAR-10}
	\label{fig:cells_cifar}
\end{figure*}

\subsection{Recurrent Neural Architecture Search} 

In the reccuent neural architecture search stage, we follow the same cell setup and search space as DARTS \cite{liu2018darts} and search for 50 epochs with batch size 256. As for the optimization, we employ AdamHAPG and the initial learning rate is set as $1 \times 10^{-4}$ and the momentum is (0.5, 0.999). The evaluation of searched architecture follows the same training strategy with DARTS but with batch size 64 and we train for 4000 epochs.

\subsection{Ablation Study}
\subsubsection{Architecture Search}
For the four competing algorithms, SGD, Adam, HAPG, AdamHAPG in the network pruning experiment, we train for 500 epochs with batch size 256 in search stage. For SGD, we employ momentum SGD with momentum 0.9 and initiate the learning rate as 0.025 with an annealing cosine decay schedule. For Adam, we initiate the learning rate as $3\times10^{-4}$ and set momentum (0.5, 0.999). For HAPG, we initiate the learning rate as 0.05 with an cosine annealing decay schedule. For AdamHAPG, the initial learning rate is set to $3\times10^{-4}$ and as for the Adam gradient update, we employ momentum (0.5, 0.999).

\subsubsection{Architecture Evaluation}
We evaluate the final sparse network after the architecture search stage. As SGD and Adam cannot directly derive a sparse network, a hard threshold is necessary and here we set the threshold as 0.001, i.e., discard the inner connections with weights smaller than 0.001 and discard neurons with all outcoming weights below 0.001. For HAPG and AdamHAPG, we simply zero out connections and neurons with zero weights. The final architecture is trained with Adam for 3000 epochs. The initiate learning rate is set to be $3\times10^{-4}$ and use momentum (0.5, 0.999).

\section{Visualization of Searched Architectures}
\begin{figure}[t]
	\centering
	% \captionsetup{font={small}}
	% \captionsetup[sub]{font=small}
	\includegraphics[width=0.5\textwidth]{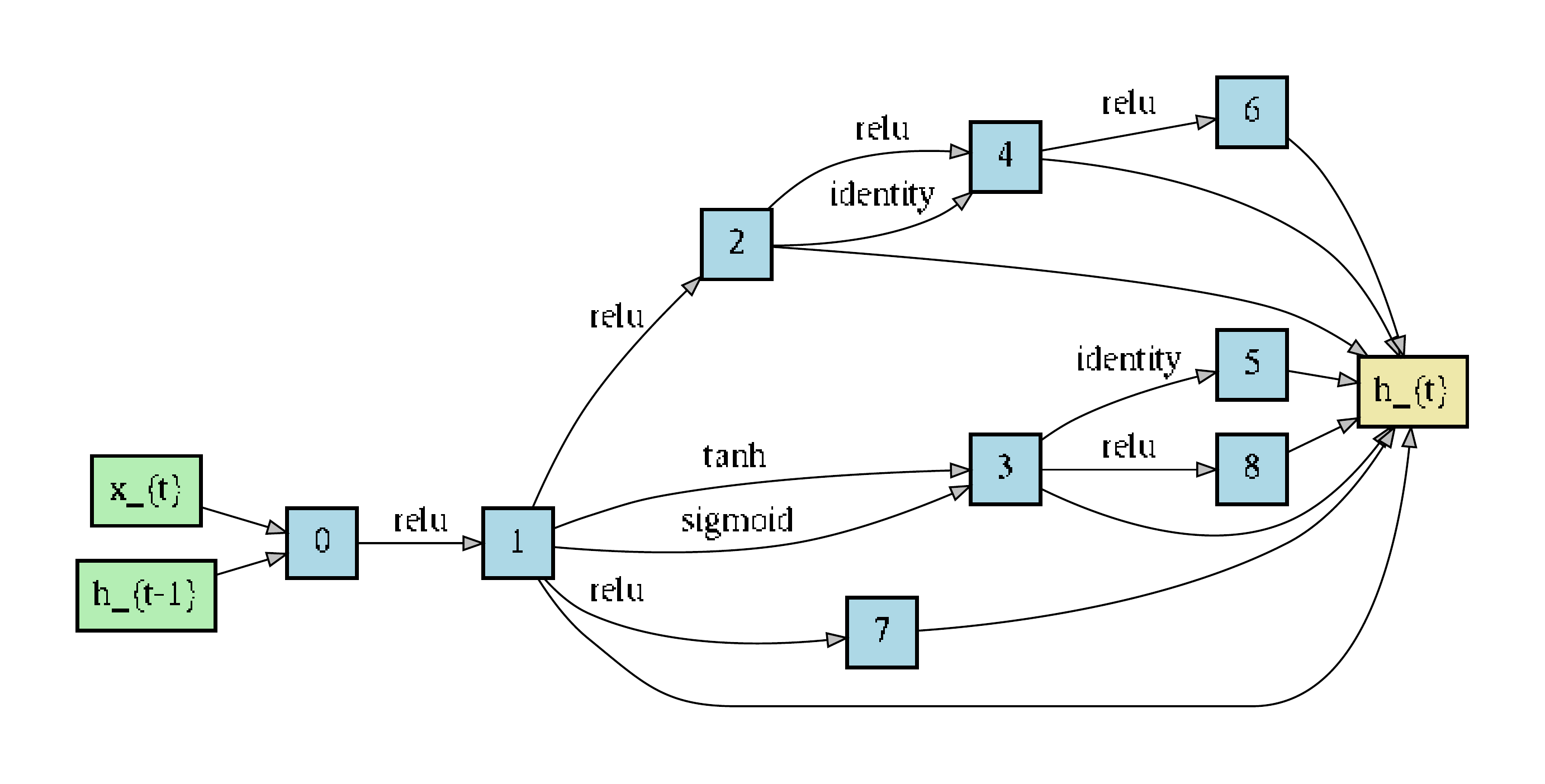}
	\caption{Recurrent cell searched on PTB}
	\label{fig:cell-ptb}
\end{figure}

We visualize the architectures searched on CIFAR-10 and PTB in Figure \ref{fig:cells-cifar} and \ref{fig:cell-ptb} . For convolutional cell, both HAPG and AdamHAPG are able to generate more general structure with flexible paths and nodes. In particular, all the resulting reduction cells are more compact with less paths and nodes. For recurrent cell, we see that there are multiple paths between some pairs of nodes, and this derived structure are shown to be effective in the major paper.

\iffalse
\begin{figure}
	\centering
	\captionsetup{font={small}}
	\captionsetup[sub]{font=small}
	\subfigure[Normal Cells]{
		\begin{minipage}[b]{0.3\textwidth}
			\centering
			\includegraphics[width=0.8\textwidth]{sparsenas/figure/normal_tiny.png}
			% \captionsetup[sub]{font=small}
			\subcaption{HAPG}
		\end{minipage}
		\begin{minipage}[b]{0.3\textwidth}
			\centering
			\includegraphics[width=1.2\textwidth]{sparsenas/figure/Tiny-adamhapg-normal.png}    \subcaption{AdamHAPG}
	\end{minipage}}
	\\
	\subfigure[Reduction Cells]{
		\begin{minipage}[b]{0.5\textwidth}
			\centering
			\includegraphics[width=0.9\textwidth]{sparsenas/figure/reduction_tiny.png}
			\subcaption{HAPG}
		\end{minipage}
		% \vspace{0.1cm}
		\begin{minipage}[b]{0.5\textwidth}
			\includegraphics[width=0.9\textwidth]{sparsenas/figure/Tiny-adamhapg-reduction.png}
			\subcaption{AdamHAPG}
	\end{minipage}}
	\caption{Architectures searched on TinyImageNet}
	\label{fig:cells-ti}
\end{figure}
\fi

\section{Comparison between Adam and the Proposed AdamHAPG}

In the main paper, we studied the superiority of our proposed optimization algorithms HAPG and AdamHAPG over the traditional SGD and Adam for the network pruning task. In this section, we further study the advantage of the proposed methods for the neural architecture search task. We compare the classification performance of Adam and the proposed AdamHAPG with different $\alpha$ values and the same $\lambda$ (i.e., $\lambda=0.01$) on CIFAR-10. As shown in table \ref{tab:alpha_ablation}, the suggested AdamHAPG algorithm can generally have clear improvement over Adam.
\begin{table}
	\centering
	\begin{tabular}{cc}
		\toprule
		& \textbf{Test Error (\%)} \\
		\midrule
		Adam~($\alpha=0.3$) & 3.32 \\
		Adam~($\alpha=0.5$) & 2.93 \\
		Adam~($\alpha=0.7$) & 3.93 \\
		
		\midrule
		AdamHAPG~($\alpha$=0.3)  & 3.30 \\
		AdamHAPG~($\alpha$=0.5)  & 2.69 \\
		AdamHAPG~($\alpha$=0.7)  & 2.79 \\
		\bottomrule
	\end{tabular}
	%    \captionsetup{font={small}}
	\caption{Comparison between Adam and the proposed AdamHAPG with various $\alpha$ values for neural architecture search on CIFAR-10}
	\label{tab:alpha_ablation}
\end{table}

% \section{Architecture Weight Distribution}

\begin{figure*}
	\centering
	\includegraphics[width=0.8\textwidth]{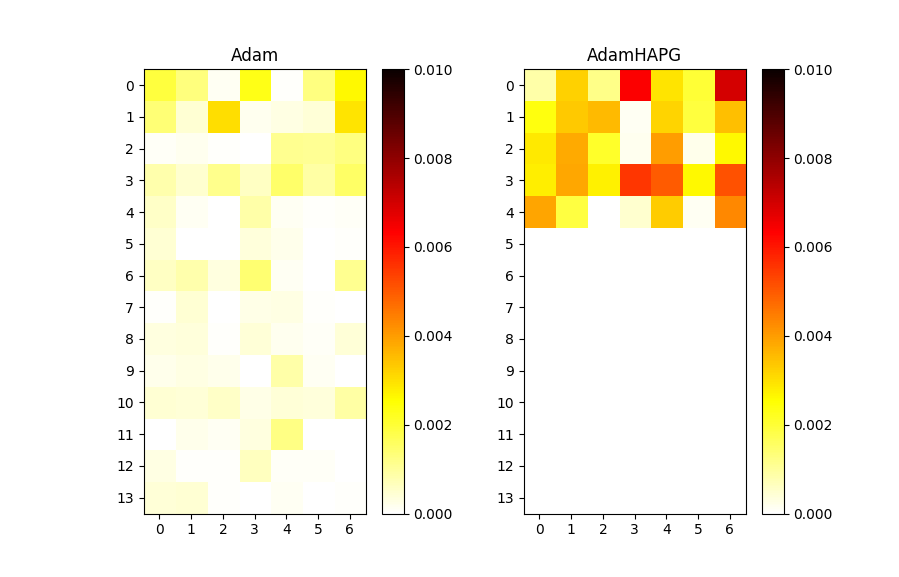}
	\caption{Architecture weight distribution of the searched cell on CIFAR-10. The horizontal axis represents the seven candidate operations, and the vertical axis indicates the input edges associated with the four nodes. The first two rows corresponds to the two input edges  associated  with  the first intermediate node, and the next three rows corresponds to the three input edges of the second intermediate node and so on. A darker color indicates a larger weight.}
	\label{fig:weights_cifar}
\end{figure*}

\begin{figure*}
	\centering
	\includegraphics[width=0.96\textwidth]{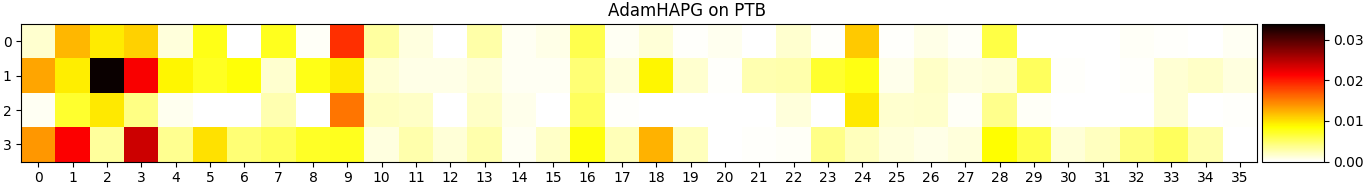}
	\caption{Architecture weight distribution of the searched cell on PTB. For the sake of presentation, each row corresponds to a candidate operation, and each column represents an input edge associated with one specific node. Particularly, the first column represents the input edge associated with the first intermediate node, and the next two columns are those of second nodes and so on. A darker color indicates a larger weight.}
	\label{fig:weights_ptb}
\end{figure*}

In addition, we compare Adam and the proposed AdamHAPG in terms of architecture weight distributions.
In Fig.\ref{fig:weights_cifar}, we visualize the weight distributions in the searched cell by Adam and AdamHAPG on the CIFAR-10 dataset. In particular, the first two rows represent the two input edges associated with the first intermediate node, and the next three rows are those of the second intermediate node and so on. Each column corresponds to a candidate operation. The darker colors indicate the larger weights. The result shows that the traditional Adam algorithm achieves very close weights for most of the candidate operations. When discarding those operations with not relatively small weights (i.e., with non-marginal contribution to the supernet), the derived final architecture may not inherit the good property of the supernet. By comparison, our AdamHAPG can result in a good weight distribution, where there are a very small number of operations with clearly bigger weights (i.e., red blocks in the weight matrix) and the removed operations generally have very small weights (i.e., most are nearly-zero) so that the derived optimal architecture can maintain the major power of the supernet. Readers can additionally find the weight distribution of  AdamHAPG on the PTB dataset in Fig.\ref{fig:weights_ptb}, which also reflects that the suggested AdamHAPG can obtain a good weight distribution.


\end{document}